\documentclass[letterpaper, 10 pt, conference]{ieeeconf}  

\usepackage[T1]{fontenc}
\usepackage[utf8]{inputenc}
\usepackage[inkscapearea=page]{svg}
\usepackage[normalem]{ulem}
\usepackage{subcaption}
\usepackage[maxfloats=256]{morefloats}
\usepackage{rotating}
\usepackage{textpos}
\usepackage{balance}
\usepackage{multirow}
\usepackage[font=scriptsize]{caption} 
\usepackage{hyperref}

\usepackage{tikz}

\usepackage[
    backend=bibtex8,
    style=ieee,
    sorting=none,
    natbib=true,
    doi=false,
    isbn=false,
    url=false,
    eprint=false,
    maxcitenames=1,
    mincitenames=1
]{biblatex}
\usepackage{acronym}
\acrodef{RMS}[RMS]{Root mean square}
\acrodef{DBH}[DBH]{Diameter at Breast Height}
\acrodef{OKS}[OKS]{Object Keypoint Similarity}
\acrodef{FCNN}[FCNN]{fully-convolutional neural network}
\acrodef{RPN}[RPN]{Region Proposal Network}
\acrodef{RoI}[RoI]{regions of interests}
\acrodef{IoU}[IoU]{intersection over union}
\acrodef{TP}[TP]{true positive}
\acrodef{FP}[FP]{false positive}
\acrodef{FN}[FN]{false negative}
\acrodef{AI}[AI]{Artificial Intelligence}
\acrodef{AP}[AP]{Average Precision}
\acrodef{AR}[AR]{Average Recall}
\acrodef{S-LOAM}[S-LOAM]{Semantic LOAM}
\acrodef{IBC}[IBC]{Intelligent Boom Control}
\acrodef{ICAS}[ICAS]{intelligent computer assisted support}
\acrodef{fps}[fps]{frame per second}
\acrodef{auc}[AUC]{area-under-curve}
\acrodef{SGD}[SGD]{stochastic gradient descent}
\acrodef{CNN}[CNN]{Convolutional Neural Network}
\acrodef{IID}[IID]{Independent and Identically Distributed}
\acrodef{FLOP}[FLOP]{Floating Point Operation}

\IEEEoverridecommandlockouts                              

\title{\LARGE \bf
Training Deep Learning Algorithms on Synthetic Forest Images for Tree Detection
}

\author{Vincent Grondin$^{1}$  $^{2}$, François Pomerleau$^{1}$ and Philippe Giguère$^{1}$
\thanks{*This work is supported by the FORAC Consortium.}
\thanks{$^{1}$ Department of Computer Science and Software Engineering 
         at Laval University, Canada }
\thanks{$^{2}$ Corresponding author: 
        {\tt\small vincent.grondin.2@ulaval.ca}}%
}

\begin{document}

\makeatletter
\let\@oldmaketitle\@maketitle
\renewcommand{\@maketitle}{\@oldmaketitle
  \includegraphics[trim=0.1cm 1cm 0.1cm 1cm, clip=true, width=1.0\linewidth]
    {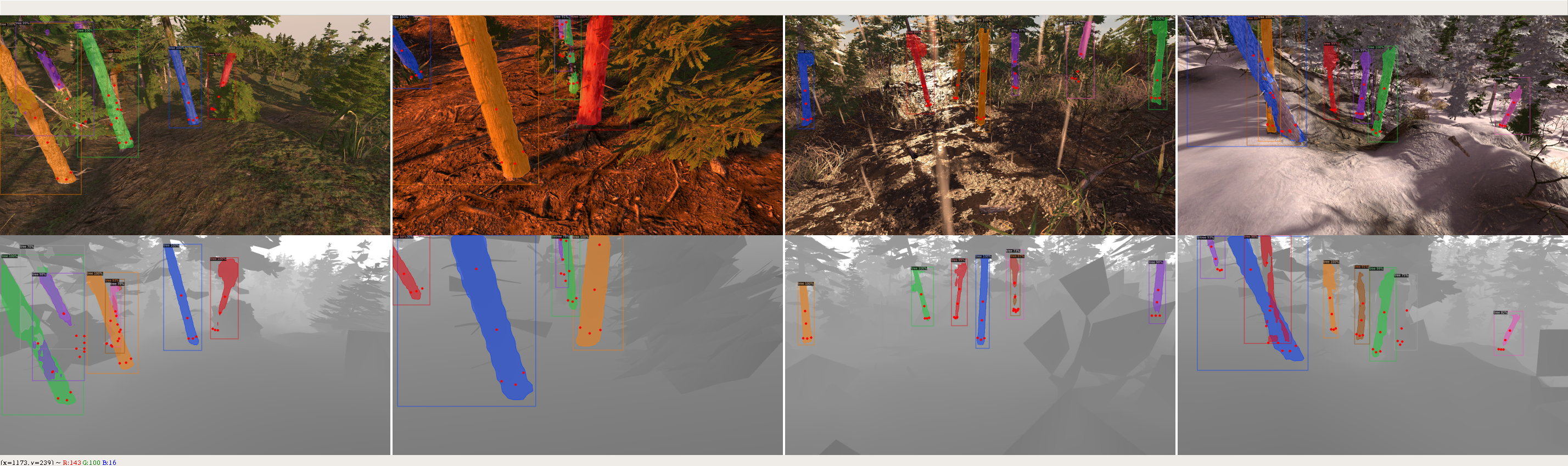} \\[\normalbaselineskip]
  \refstepcounter{figure}\centering\scriptsize{Fig.~\thefigure: Predictions by \texttt{ResNeXt-101} on synthetic RGB images (top row) and by \texttt{ResNet-101} on depth images (bottom row). The models achieve qualitatively good results under challenging simulated conditions such as light variation, occlusion, instances overlapping and weather effects.}
  \label{fig:synth_detections}
}
\makeatother

\maketitle
\thispagestyle{empty}
\pagestyle{empty}


\begin{abstract}

Vision-based segmentation in forested environments is a key functionality for autonomous forestry operations such as tree felling and forwarding. 
Deep learning algorithms demonstrate promising results to perform visual tasks such as object detection. 
However, the supervised learning process of these algorithms requires annotations from a large diversity of images.
In this work, we propose to use simulated forest environments to automatically generate 43\,k realistic synthetic images with pixel-level annotations, and use it to train deep learning algorithms for tree detection.
This allows us to address the following questions: i) what kind of performance should we expect from deep learning in harsh synthetic forest environments, ii) which annotations are the most important for training, and iii) what modality should be used between RGB and depth. 
We also report the promising transfer learning capability of features learned on our synthetic dataset by directly predicting bounding box, segmentation masks and keypoints on real images. Code available on GitHub (https://github.com/norlab-ulaval/PercepTreeV1).  

\end{abstract}

\section{Introduction}
\label{sec:intro}

Deep learning gained much attention in the field of forestry as it can implement knowledge into machines to tackle problems such as tree detection or tree health/species classification~\cite{diez2021deep}. 
However, deep learning is a data centric approach that needs a sufficient amount of annotated images to learn  distinctive object features.
Creating an image dataset is a cumbersome process requiring a great deal of time and human resources, especially for pixel-level annotations.
Accordingly, few datasets specific to forestry exist, and this limits deep learning applications, as well as task automation requiring high-level cognition.

In order to avoid hand-annotation and include as many realistic conditions as possible in images, we propose to fill the data gap by creating a large dataset of synthetic images containing over 43\,k images, which we name the \textsc{SynthTree43k} dataset.
Based on this dataset, we train Mask R-CNN~\cite{he2017mask}, the most commonly-used model for instance segmentation~\cite{diez2021deep}, and measure its performances for tree detection and segmentation. 
Because our simulator allows for quick annotation, we also experiment with keypoint detection to provide information about tree diameter, inclination and felling cut location.  

Even though the detection performances obtained on \textsc{SynthTree43k} will not directly transfer to real world images because of the reality gap, a result analysis can guide us towards building an optimal real dataset.
Notably, synthetic datasets can be used to evaluate preliminary prototypes~\cite{gaidon2016virtual}, and sometimes they can improve detection performance when combined with real-world datasets~\cite{gaidon2016virtual, de2017procedural}.
In that sense, we shed light on which annotations are the most impactful on learning, and if adding the depth modality in the dataset is pertinent.
Lastly, we demonstrate the reality gap by qualitatively testing the model on real images, showing transfer learning potential.

\section{Related Work}
\label{sec:sota}
Deep learning for tree detection in forestry has demonstrated success on relatively small real image datasets. 
For instance, \cite{liu2019classification} implement a U-Net architecture to perform tree specie classification, detection, segmentation and stock volume estimation on trees. 
When trained on their (private) dataset of 3\,k images, they achieve 97.25\,\% precision and 95.68\,\% recall rates. 
Similarly, \cite{da2021visible} uses a mix of visible and thermal images to create a dataset of 2895 images extracted from video sequences, and solely include bounding box annotations.
They trained five different one-shot detectors on their dataset and achieved 89.84\,\% precision, and 89.37\,\% F1-score. 

We believe these aforementioned methods could benefit from training on synthetic images. 
The Virtual KITTI dataset~\cite{gaidon2016virtual} is one of the first to explore this approach to train and evaluate models for autonomous driving applications. 
By recreating real-world videos with a game engine, they generate synthetic data comparable to real data.
The models trained on their virtual dataset show that the gap between real and virtual data is small, and it can substitute for data gaps in multi-object tracking.
Meanwhile, ~\cite{ros2016synthia} explore object detection using a synthetic dataset for autonomous driving, and they report that training models on realistically rendered images could produce good segmentations by themselves on real datasets while dramatically increasing accuracy when combined with real data. 
Quantitatively, they improved per-class accuracy by more than 10 points (and in some cases, as far as  18.3 points).

Although synthetic datasets cannot completely replace real world data,  multiple works demonstrate that it is a cost-effective alternative that offers good transferability  ~\cite{gaidon2016virtual, ros2016synthia, de2017procedural}.
Therefore, developing synthetic forest datasets will potentially improve the current state of tree perception methods in forestry.

\section{Methodology}

In this section, we detail how \textsc{SynthTree43k} was created.
Then, we describe the deep learning architecture and backbones, as well as the training details.

\subsection{Simulator and Dataset} 
\textsc{SynthTree43k} is generated by employing the Unity\footnote{https://unity.com} game engine to render realistic virtual forests.
This virtual world generator can be configured via Gaia\footnote{https://assetstore.unity.com/packages/tools/terrain/gaia-2-terrain-scene-generator-42618} to procedurally terra-form the landscape, texture the terrain and spawn objects.
In this simulation, the forest density is controlled through various spawn rules such as altitude, terrain slope and the number of neighbouring objects in a given area. 

\begin{figure}[htbp]
    \centering
    \includegraphics[width=1.0\linewidth]{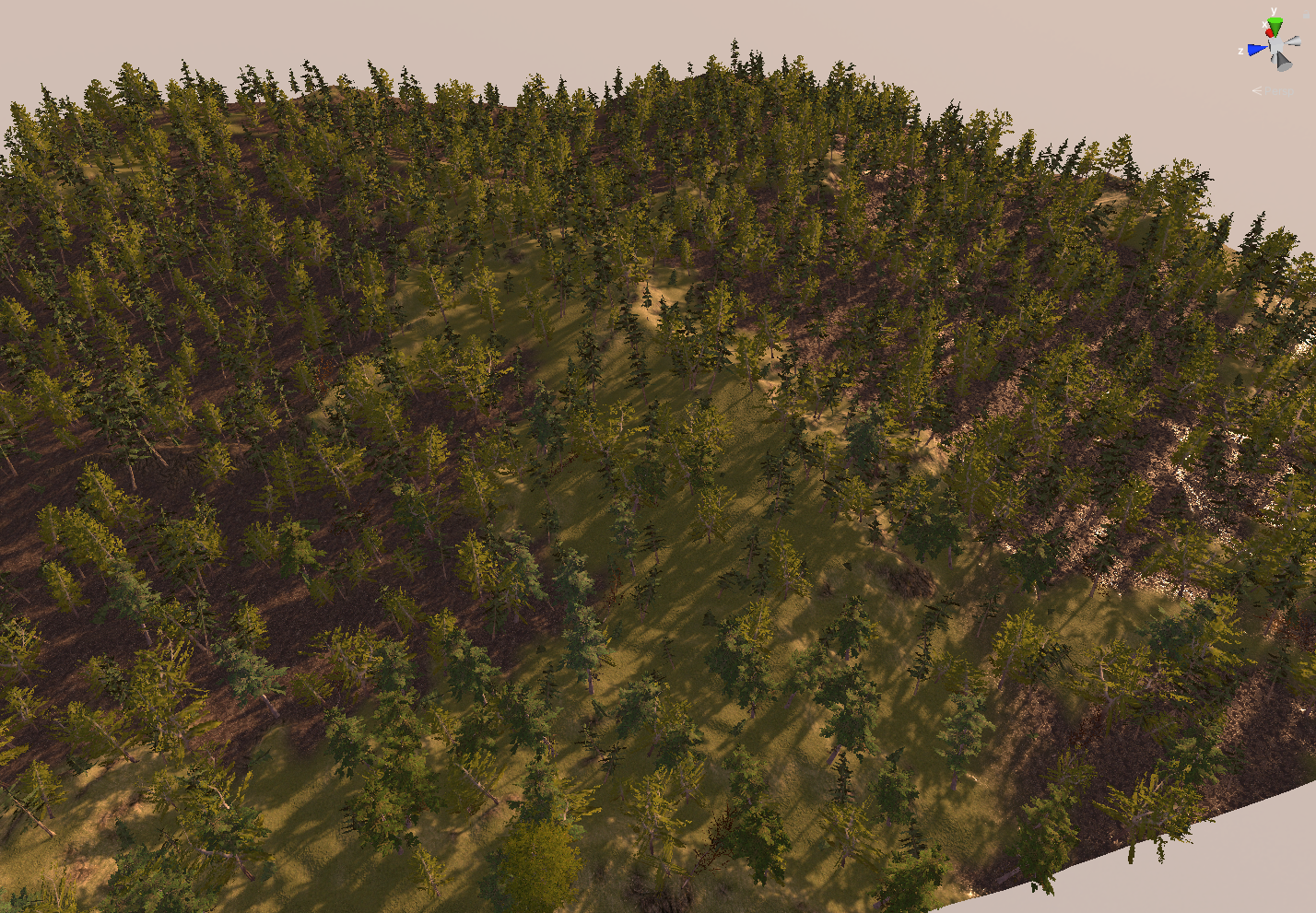}
    \caption{A general view of a simulated forest environment. In this scene, three terrain textures are used to simulate moss, roots and mud conditions. The tree models are fir and beech, accompanied by scrubs, branches, grass and stomps under a morning light effect. }
    \label{fig:keypoints_hist}
\end{figure}

The forest is populated with realistic tree models from Nature Manufacture\footnote{https://naturemanufacture.com/}.
In order to extend visual variability, texture on the six tree models is modified to create 17 new, distinctive tree models. 
Other object models from Nature Manufacture are also included in scenes such as grass, stumps, scrubs and branches.

For additional realism and variety, meteorological conditions are also simulated in this virtual world. 
We simulate snow using snow texture, and wet effect using decals. 
Particle systems are employed to recreate snowflakes, raindrops or fog effects.  
To simulate different moments of the day, we adjust illumination to morning, daylight, evening and dusk. 
The object shadows adapt to illumination cast on the scene.

From each generated scene, we add between 200-1000 images to the dataset.
Each image includes bounding box, segmentation mask and keypoint annotations.
The pipeline annotates trees within a 10\,m radius from the camera, which corresponds to the reach of a harvester~\cite{lindroos2015estimating}.
Five keypoints are assigned per trees to capture the essential information that an autonomous tree felling system would need: the felling cut location,
diameter and inclination. 

Collectively, this pipeline can generate an unlimited amount of synthetic
images with an annotation speed of approximately 20 frames/minute, for we consider
RGB and depth images as one frame. 
Overall, \textsc{SynthTree43k} contains over 43\,k RGB and depth images, and over 162\,k annotated trees.

\subsection{Models}
The Mask R-CNN architecture is composed of \emph{i)} a convolutional feature extraction backbone, \emph{ii)} a \ac{RPN}, and \emph{iii)} prediction heads.
The original Mask R-CNN network is slightly adapted for our tree detection problem, in that, an optional keypoint branch is added to the prediction head. Therefore, it can be used for classification, bounding box regression, segmentation, and keypoint prediction. 

Predictions are made via a two-stage process.
In the first stage, the \ac{RPN} proposes \ac{RoI} from the feature maps of the backbone. 
These correspond to a region that potentially contains a tree. 
The generation of a \ac{RoI} follows along the default nine box anchors, corresponding to three area-scales (8, 16 and 32) and three aspect ratios (0.5, 1.0, and 2.0). 
In our experiment, we employ three different backbone architectures: ResNet-50, ResNet-101~\cite{he2016deep}, and ResNeXt-101~\cite{xie2017aggregated}. 
We use ResNet backbone for feature extraction as it gives excellent gains in both accuracy and speed, which we use as a baseline for our results.
When comparing the 50-layer to its 101-layer counterpart, there is a possibility that they perform similarly when the dataset is small~\cite{he2017mask}, yet we expect that using \textsc{SynthTree43k} will be enough to demonstrate that the 101-layer can outperform the 50-layer.
In regards to ResNeXt, it introduces a cardinality hyper-parameter, which is the number of independent paths, providing a way to adjust the model capacity without going deeper or wider.
More details about backbone parameters are provided in \autoref{tab:backbones}

\begin{table}[h]
\caption{Backbone parameters. The number of learnable parameters (\#Params), computational complexity (GFLOPs) and frames per second (FPS) at inference time on $800\times800$ images.}
\label{tab:backbones}
\begin{center}
\begin{tabular}{|c|ccc|}
    \hline
    Backbone & \#Params & GFLOPs & FPS \\
    \hline
    \texttt{ResNet-50-FPN} & 25.6\,M & 3.86 & 18\\
    \hline
    \texttt{ResNet-101-FPN} & 44.7\,M & 7.58 & 15 \\
    \hline
    \texttt{ResNeXt-101-FPN} & 44\,M & 7.99  & 10 \\
    \hline
\end{tabular}
\end{center}
\end{table}

Subsequently, RoIAlign~\cite{he2017mask} uses bilinear interpolation to map the feature maps of the backbone into a $7\times7$ input feature map within each \ac{RoI} area. Features from each \ac{RoI} then go through the network head to simultaneously predict the class, box offset, binary segmentation mask, and an optional binary mask for each keypoint.

\subsection{Training Details}

We use Detectron2~\cite{wu2019detectron2} implementations of Mask R-CNN.
It has been shown that pre-training helps regularize models~\cite{erhan2010does}, and facilitate transfer learning to a target domain~\cite{mahajan2018exploring}.
Therefore, the Mask R-CNN models employed in our experiments are pre-trained on the COCO Person Keypoint dataset~\cite{lin2014microsoft}, which is a large-scale dataset containing more than 200\,k images and 250\,k person instances labeled with 17 keypoints per instance. 
Before training or fine-tuning the models, the first two convolutional layers of the backbone are frozen. 
The hardware for model training and testing is an NVIDIA RTX-3090-24GB GPU and an Intel Core i9-10900KF CPU.

To train the model, \textsc{SynthTree43k} is split into three subsets: 40\,k in the train set, 1\,k for the validation set and 2\,k in the test set. 
The model learns from the train set by using an \ac{SGD} optimizer with a momentum of 0.9, and a weight decay of 0.0005. 
During training, we improve model generalization, and reduce dataset overfitting by employing data augmentation techniques such as image resizing, horizontal flipping, sheering, saturation, rotation, and cropping. 
No data augmentations are used at validation and test time. 
Model overfitting is monitored via the validation set, which is also used for early stopping.

Depth images are gray scales of 8-bit 1-channel.
At train time, they are converted to 8-bit 3-channel to fit the RGB format from pre-trained models. 
In our case, a single image channel could be possible, but it would require randomly initiated models, which takes an enormous amount of pre-training time for backbones such as \texttt{ResNet} and \texttt{ResNeXt}. 

Hyperparameter optimization is conducted for \texttt{ResNet-50-FPN} only, and these hyperparameters are used for every model. 
We use early stopping based on the highest validation set \ac{AP} to determine when to stop training.

\section{Experimental Results}
We base our performance evaluation on the standard COCO metrics for each task, AP$^{bb}$ and AP$^{mask}$, we train Mask R-CNN on our synthetic forest images and compare detection performance between the three backbones and RGB/Depth modality. 
We also conduct an analysis of keypoint prediction by measuring the pixel error of each predicted keypoint. 
Lastly, we test detection on real images, qualitatively demonstrating the reality gap between synthetic and real images. 

\begin{figure*}[htbp]
    \centering
    \includegraphics[width=1.0\linewidth]{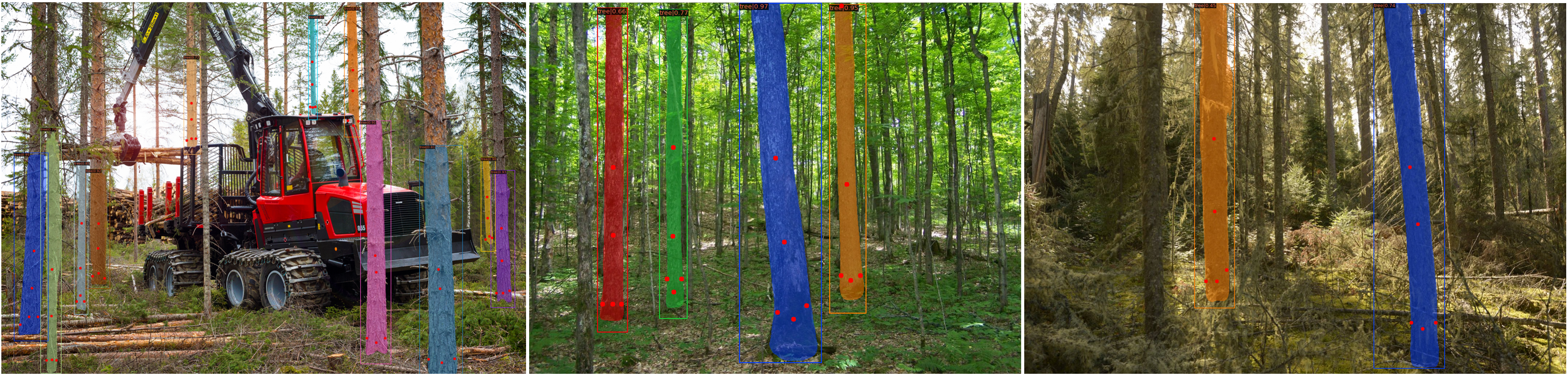}
    \caption{Predictions on real images from \texttt{ResNeXt-101} trained only on synthetic images. We observe that even with the reality gap, the model can still detect trees with high precision, but suffers from low recall rates. }
    \label{fig:pred_on_real}
\end{figure*}

\subsection{Tree Detection and Segmentation}
\label{subsec:tree_detection}
Six models, corresponding to all combinations between the three different backbones and two modalities, are trained and tested on \textsc{SynthTree43k}.
From \autoref{tab:AP_synth}, we observe that all models trained on the depth modality outperform models trained on the RGB modality.
In fact, the detection task based on the depth modality improves AP$^{bb}$ by 9.49\,\%  on average, even though all of the models were pre-trained on COCO Person --- an RGB dataset. 
This suggests two things: 1) that the depth modality helps networks reject trees located further than our 10\,m annotation threshold, and 2) depth images are possibly easier to interpret. 
In regard to the segmentation task, very little gain is obtained by using depth images instead of RGB, and in the case of \texttt{ResNeXt-101} it decreases.
Surprisingly, the \texttt{ResNeXt} architecture has more trouble transferring to depth images compared to the \texttt{ResNet} architecture, which make \texttt{ResNet-101} the best backbone for depth. 

On RGB images, we achieve the best detection results using \texttt{ResNeXt-101}.
This is similar to previous research~\cite{xie2017aggregated, wu2020object}, and it is a result of the cardinality used in \texttt{ResNeXt} as it is more effective than going deeper or wider when the model capacity is increased.
The predictions on synthetic images can be observed in \autoref{fig:synth_detections}.

\begin{table}[htbp]
\caption{Results for models trained and tested on \textsc{\textsc{SynthTree43k}}. All models achieved better performances using the depth modality.}
\label{tab:AP_synth}
\centering
\begin{tabular}{c|c|cc|cc}
    Backbone & Modality & AP$^{bb}$ & AP$^{mask}$ & AP50$^{bb}$ & AP50$^{mask}$\\ 
    \hline
    \multirow{2}{*}{\texttt{R-50}} & RGB & 55.20 & 31.13  & 87.74 & 69.36\\ 
    & Depth & 66.70 & 31.52 & 89.67 & 70.66\\ 
    \hline
    \multirow{2}{*}{\texttt{R-101}} & RGB & 56.79 & 31.72 & 88.51 & 70.53 \\ 
    & Depth & \textbf{68.20} & \textbf{31.98} & \textbf{89.89} & \textbf{71.65}\\ 
    \hline
    \multirow{2}{*}{\texttt{X-101}} & RGB & 58.34 & 31.77 & 88.91 & 71.07 \\ 
    & Depth  & 63.90 &	28.86 & 87.41 & 68.19 \\ 
\end{tabular}
\end{table}

\autoref{tab:multi-task_learning} shows that adding the mask branch consistently improves AP$^{bb}$ and AP$^{kp}$.
Comparatively, adding the keypoint branch reduces AP$^{bb}$ and AP$^{mask}$.
These findings align with \cite{he2017mask}, as they found that the keypoint branch benefits from multitask training, but it does not help the other tasks in return. 
Better bounding box and keypoint detections can occur by learning the features specific to segmentation.
In fact, a richer and detailed understanding of image content requires pixel-level segmentation, which can play an important role in precisely delimiting the boundaries of individual trees~\cite{liu2020deep}.

\begin{table}[ht]
\caption{Impact of multi-task learning on bounding box, segmentation mask, and keypoints. Results are from \texttt{ResNeXt-101} on real RGB images. 
}
\label{tab:multi-task_learning}
\centering
\renewcommand{\arraystretch}{1.1}
\begin{tabular}{c|ccc}
    Tasks & AP$^{bb}$ & AP$^{mask}$ & AP$^{kp}$ \\ 
    \hline
    mask-only & \textbf{59.25} & \textbf{32.65} & -   \\ 
    \cline{1-4}
    keypoint \& mask & 58.34 & 31.77 & \textbf{80.19}  \\ 
    \cline{1-4}
    keypoint-only & 57.71 & - & 80.13  \\ 
    
\end{tabular}
\end{table}

\subsection{Keypoint Detection}
\label{subsection:keypoint_detection}
A keypoint detection analysis based on error in pixels allows for meaningful and straightforward interpretations of tree-felling tasks and their error distribution.
Therefore, we compute the pixel error between the ground truth and the predicted keypoint, and report the results in \autoref{fig:keypoints_hist}. 

\begin{figure}[htbp]
    \centering
    \includegraphics[width=1.0\linewidth]{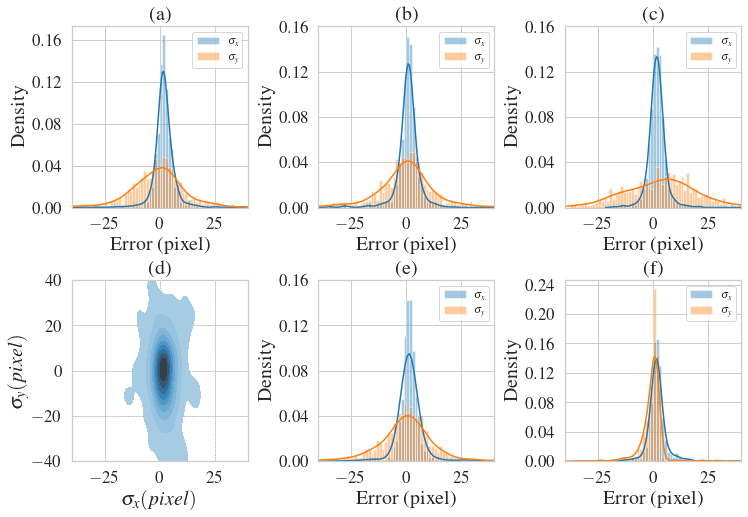}
    \caption{Keypoint error distributions (in pixels) for our best detection model, \texttt{ResNet-101} on depth images. Blue is the horizontal error and orange is the vertical error distribution for the \textbf{(a)} felling cut keypoint, \textbf{(b)} right and \textbf{(e)} left diameter keypoint; \textbf{(c)} middle keypoint and \textbf{(f)} top keypoint. Density map of the felling cut keypoint is shown in \textbf{(d)}. }
    \label{fig:keypoints_hist}
\end{figure}

Our tree detection model achieves a mean error of 5.2\,pixels when estimating tree diameters.
This accuracy is promising for automation applications, depending on the distance between the tree and the camera.
Since the conversion from pixel error to metric error depends on the depth, the error increases when the tree is further away, and in turn decreases accuracy.

We observe a significant difference between horizontal and vertical error, where $\sigma_y$ is about three times the value of $\sigma_x$, for the felling cut, right and left diameter keypoint, and middle keypoint.
We expected a larger $\sigma_y$ than $\sigma_x$, because the horizontal keypoint position is either located on the side or center of the trees.
In comparison, the vertical position of each keypoint is subjectively more difficult to estimate due to the inability of extracting precise vertical information.
Hence, the difficulty in estimating the $\sigma_y$ error greatly impacts the estimated felling cut position.
If it is estimated too high on the stem, felling the tree will leave behind high stumps that are against current harvesting practices \cite{ireland2009CCF}. 
Keypoint predictions with high vertical values often occur when a dense understorey restricts the line of sight to the tree base. 
This causes faulty predictions to position above the understorey, which results in an inappropriate felling cut height.
In practice, a simple solution to this issue is to place the felling head on the predicted point, and roll it down to the base~\cite{ireland2009CCF}.

\subsection{Prediction on Real Images}
\label{subsection:pred_real_images}
We test the transferability of our model on real images.
Due to the lack of real image datasets for tree detection and segmentation, the models are not fine-tuned on real images.
Qualitative results can be observed in \autoref{fig:pred_on_real}.
Visually, we see that not only bounding boxes are well predicted, but segmentation masks along with keypoint predictions are also transferred successfully.
The model seems to be more precise than accurate, which indicates that it is unable to detect trees that are too different from the ones trained on in the synthetic dataset. 
Adding different tree models to our simulation could help generalize to the real world.
Virtual pre-training is a promising practice given the current data gap in forestry compared to other domains, like autonomous driving or industrial automation.

\section{Conclusion and Future Works}

In short, we explored the use of synthetic images to train deep learning algorithms for tree detection. 
We provide quantitative experimental evidence suggesting that the segmentation task is important and helps to improve both bounding box and keypoint predictions.
Therefore, the creation of a real image dataset in forestry should include these annotations. 
We also show that the depth modality significantly outperforms the RGB modality in the synthetic world.
Finally, we qualitatively demonstrate that direct transfer to real world images suffer from low accuracy, while the precision is relatively good.
Models publicly available\footnote{https://github.com/norlab-ulaval/PercepTreeV1}.

In future works, we plan to evaluate tree detection  performances on a real images dataset and assess its possible use in forestry related operations.


\printbibliography

\end{document}